\documentclass{bmvc2k}
\usepackage{booktabs}
\usepackage{graphicx}
\usepackage{multirow}
\usepackage[table,xcdraw]{xcolor}
\usepackage{amssymb}
\usepackage{comment}
\usepackage{siunitx}

\usepackage[symbol]{footmisc}
\renewcommand{\thefootnote}{\fnsymbol{footnote}}

\title{DKMA-ULD: Domain Knowledge augmented Multi-head Attention based Robust Universal Lesion Detection}

\addauthor{Manu Sheoran*}{manu.sheoran@tcs.com}{1}
\addauthor{Meghal Dani*}{dani.meghal@tcs.com}{1}
\addauthor{Monika Sharma}{monika.sharma1@tcs.com}{1}
\addauthor{Lovekesh Vig}{lovekesh.vig@tcs.com}{1}

\addinstitution{
 TCS Research\\
 New Delhi, India
}

\runninghead{DKMA-ULD}{Robust Universal Lesion Detector}

\newcommand\meghal[1]{\textcolor{orange}{[Meghal: #1]}}

\begin{document}

\maketitle
\begin{abstract}
\let\thefootnote\relax\footnote{* Equal contribution in paper}
Incorporating data-specific domain knowledge in deep networks explicitly can provide important cues beneficial for lesion detection and can mitigate the need for diverse heterogeneous datasets for learning robust detectors. In this paper, we exploit the domain information present in computed tomography (CT) scans and propose a robust universal lesion detection (ULD) network that can detect lesions across all organs of the body by training on a single dataset, DeepLesion. We analyze CT-slices of varying intensities, generated using heuristically determined Hounsfield Unit (HU) windows that individually highlight different organs and are given as inputs to the deep network. The features obtained from the multiple intensity images are fused using a novel convolution augmented multi-head self-attention module and subsequently, passed to a Region Proposal Network (RPN) for lesion detection. In addition, we observed that traditional anchor boxes used in RPN for natural images are not suitable for lesion sizes often found in medical images. Therefore, we propose to use lesion-specific anchor sizes and ratios in the RPN for improving the detection performance. We use self-supervision to initialize weights of our network on the DeepLesion dataset to further imbibe domain knowledge. Our proposed Domain Knowledge augmented Multi-head Attention based Universal Lesion Detection Network \textbf{DMKA-ULD} produces refined and precise bounding boxes around lesions across different organs. We evaluate the efficacy of our network on the publicly available DeepLesion dataset which comprises of approximately $32K$ CT scans with annotated lesions across all organs of the body. Results demonstrate that we outperform existing state-of-the-art methods achieving an overall sensitivity of $87.16\%$.
\end{abstract}

\section{Introduction}
\label{sec:intro}
Advances in deep learning techniques have led to significant breakthroughs in medical image analysis~\cite{litjens2017survey, meta-dermdiagnosis, ronneberger2015u, singh20203d}. In the past, efforts have been made to build automated lesion detection solutions that focus on specific organs such as liver, kidney, and lungs~\cite{liver_detect, kidney, pulm_nodule}. However, to address the clinical necessity where radiologists are required to locate different types of lesions present in various organs of the body to diagnose patients and determine treatment, developing a universal lesion detection (ULD)~\cite{zlocha2019universal, tang2019uldor, yan2019mulan, li2019mvp, 3dce} model has become an active area of research. Tang et. al~\cite{tang2019uldor} proposed ULDor based on Mask-RCNN for lesion detection and a hard negative mining (HNEM) strategy to reduce false positives. However, the proposed HNEM technique may not enhance detection performance due to missing annotations as the mined negatives may actually contain positives. There are a few more impressive RCNN based ULD networks that use weights pre-trained on Imagenet for detection~\cite{riblidetect, 3dce, yan2019mulan}. We also found from earlier methods~\cite{volumetric, 3dce} that utilizing neighboring slice information is essential for providing $3$D context to the network which gives a lift in lesion detection accuracy. This is due to the fact that clinicians look at multiple slices of a patient's CT scan to confirm the final diagnosis. 

There also exist attention-based ULD networks where attention has been shown to improve the lesion detection~\cite{li2019mvp,3dce_improving, volumetric,retinanet_improv} by enabling the network to focus on important regions of CT-scans. MVP-Net~\cite{li2019mvp} proposed to use a position-aware attention module to aggregate features from a multi-view feature pyramid network. Another work on ULD by Wang et al.~\cite{volumetric}, proposed volumetric attention which exploits $3$D-context from multi-slice image inputs and a $2.5$D network for improving the detection performance. The multi-task universal lesion analysis network (MULAN)~\cite{yan2019mulan} utilizes $27$ slices as input and proposes a $3$D feature fusion strategy with Mask-RCNN backbone for lesion detection. In addition, they jointly train the network to perform lesion segmentation and tagging.

Typically, deep networks are reliant on high-volume datasets for automatically discovering relevant features for a learning task. However, due to the very similar appearance of lesions and other internal structures in CT scans, lesion detection is quite a challenging problem. Yan et al.~\cite{yan2020universal} proposed a Lesion ENSemble (LENS) network for lesion detection that can efficiently learn from heterogeneous lesion datasets and address the issue of missing annotations by exploiting clinical prior knowledge and cross-dataset knowledge transfer. In another paper~\cite{meld}, authors have proposed a MELD network for lesion detection which learns from multiple heterogeneous diverse datasets and uses missing annotation matching (MAM) and negative region mining (NRM) for achieving state-of-the-art lesion detection performance on DeepLesion~\cite{yan2018deeplesion} dataset. In summary, previous works on ULD have made use of 3D-context in the form of multi-slice inputs, incorporation of attention mechanisms, multi-task learning, hard negative mining techniques, and multiple heterogeneous datasets to enhance the lesion detection sensitivity performance. 

Rather than using a variety of heterogeneous datasets to learn robust representations for ULD, we claim that significant improvements to learning can be obtained by incorporating task-specific domain knowledge in the network. This motivates us to extract as many domain-specific features as possible from a minimal number of CT-slices for a particular patient, to come up with a computationally efficient ULD with enhanced prediction performance. Therefore, in our proposed lesion detector, we utilize only $3$ slices from a patient's CT scan to incorporate 3D context in the network. Next, taking cues from MVP-Net~\cite{li2019mvp}, we utilize the information of tissue density from CT-scans represented as HU-values in terms of window width and window length. During manual analysis, radiologists adjust these windows to focus on organs/tissues of interest~\cite{bae2005ct}. Despite having critical importance in lesion detection, the use of multiple HU windows of CT-slices, representing different organs of the body, has been overlooked in the literature. To this end, we introduce $5$ novel heuristically determined HU windows for CT-slices and feed them as multi-intensity input to the detection network to make it organ-agnostic. Further, the computed features are combined using our novel convolution augmented multi-head attention-based fusion architecture. We use transformers~\cite{attentionisallyouneed, ViT, detr} based self-attention mechanism for feature-fusion of multi-intensity images that effectively combines multi-organ information efficiently. This is analogous to the radiologists' way of paying attention to different organs at different HU windows of CT-slices simultaneously while detecting lesions. Additionally, we have observed that default anchor sizes and ratios used in general object detection networks do not perform satisfactorily for lesions of different sizes, particularly for very small ($<10$mm) and medium-sized ($10-30$mm) lesions. Therefore, we propose new anchor sizes and ratios for RPN that enable the network to detect lesions of varied sizes mostly found in medical imaging datasets. We named our network \textbf{\emph{DKMA-ULD}} (Domain-knowledge augmented multi-attention based universal lesion detection).

\begin{figure*}[t]
  \centering
  \includegraphics[width=0.85\linewidth]{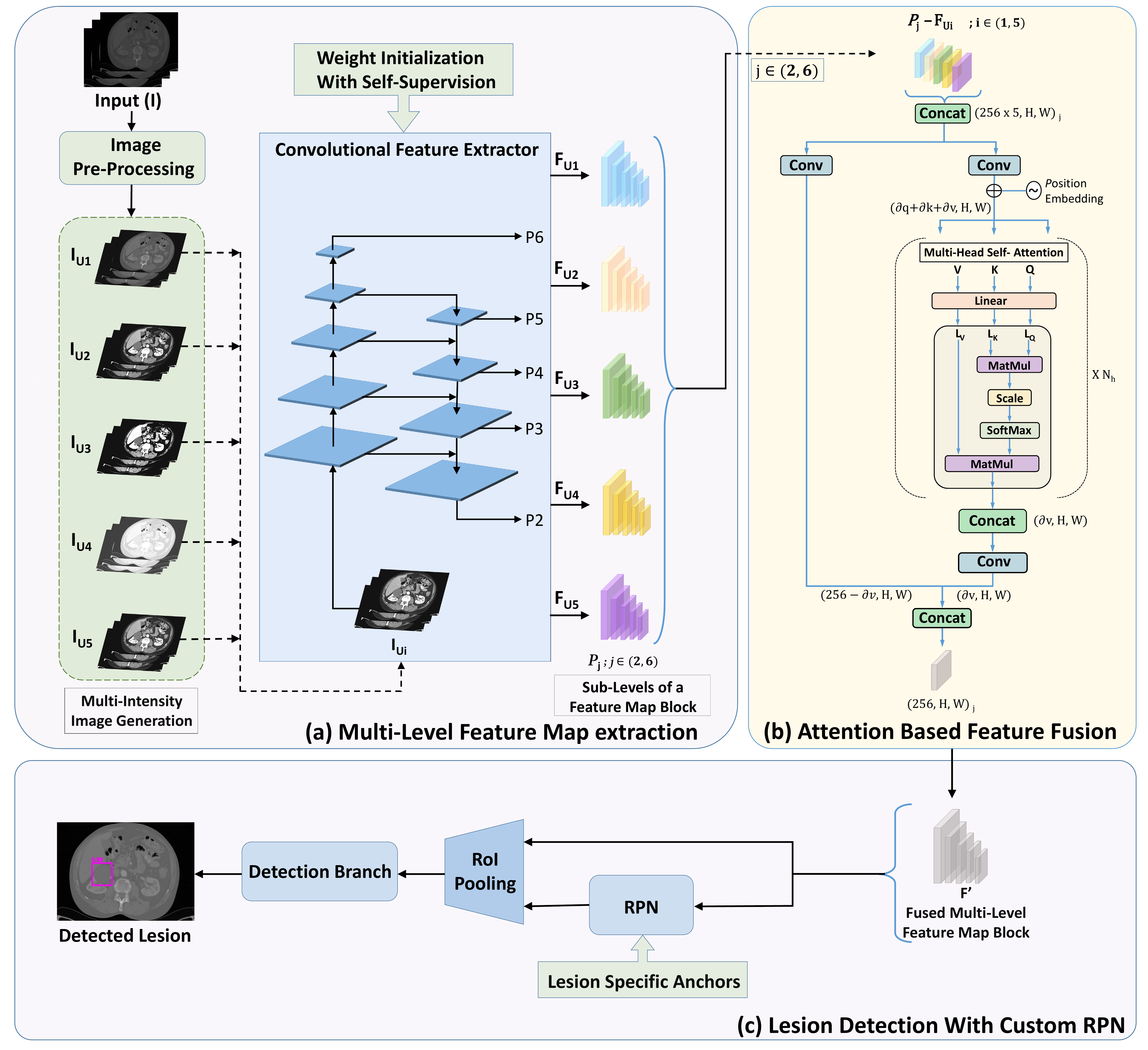}
  \caption{\small{Overview of our \emph{DKMA-ULD} architecture. (a) An input image $I$, consisting of $3$ CT-slices of a patient, is pre-processed and used for generating $5$ multi-intensity images using $5$ different HU windows (${U_{1}}$ to ${U_{5}}$). Using a shared convolutional feature extractor (having domain-specific weight initialization) with $5$ FPN levels (${P_{2}}$ to ${P_{6}}$), we obtain $5$ feature map blocks ${F_{Ui}}$, (b) Using our proposed attention based feature fusion module inspired from transformer's multi-head self-attention~\cite{attentionisallyouneed}, feature map sub-levels $({P_{j}}-{F_{Ui}})$ with same resolution of $(H, W)$ are fused into a single sub-level for obtaining the final fused feature map block (F$'$). Here V, K, Q and $N_h$ represent value, key, query matrix and number of attention heads and finally, (c) RPN with custom lesion-specific anchors is applied over F$'$ for improved lesion detection.}}
  \label{fig:arch-dia}
  \vspace{-3mm}
\end{figure*}

Furthermore, observing that self-supervised learning (SSL) techniques have recently become the cornerstone for learning improved representations in data-scarce scenarios which can, subsequently, be used for efficient learning on downstream tasks~\cite{simclr, ssl, grill2020bootstrap}. We utilize Bootstrap Your Own Latent (BYOL)\cite{grill2020bootstrap}, a SSL technique to learn weights of the backbone network of \emph{DKMA-ULD} on the DeepLesion~\cite{yan2018deeplesion} dataset. The DeepLesion dataset consists of approximately $32K$ CT scans with annotated lesions across different organs of the body. Subsequently, the weights learned via SSL are used for initializing our \emph{DKMA-ULD} so that it is able to learn robust domain-specific representations resulting in improved sensitivity. To summarize, we make the following contributions in the paper: 
\begin{itemize}
    \vspace{-2mm}
    \item We propose a domain-knowledge augmented deep network with multi-head self-attention based feature-fusion named \emph{DKMA-ULD} which performs robust detection of lesions across all organs of the body and is trained on DeepLesion~\cite{yan2018deeplesion} dataset.
    \vspace{-2mm}
    \item We introduce $5$ novel HU windows, computed in an unsupervised manner, for highlighting the different organs of the body in CT -scans which make our network organ-agnostic. 
    \vspace{-2mm}
    \item We propose a novel convolution augmented multi-head self-attention mechanism for the fusion of the features obtained from multiple intensity CT-slices for subsequent detection by an RPN.
    \vspace{-2mm}
    \item We propose lesion-specific new anchor sizes and ratios for detection that cover various sizes of lesions present in medical images. We also illustrate that these new anchors can detect very small and medium-sized lesions effectively. Hence, giving a boost to the overall detection sensitivity.
    \vspace{-2mm}
    \item We demonstrate that initializing DKMA-ULD with self-supervised domain-specific weights is beneficial for learning improved representations over Imagenet weights.
    \vspace{-2mm}
    \item We evaluate \emph{DKMA-ULD} on DeepLesion dataset and show improvement over existing state-of-the-methods of ULD such as MULAN~\cite{yan2019mulan}, 3DCE~\cite{3dce}, MVP-Net~\cite{li2019mvp}, MELD~\cite{meld} and improved RetinaNet~\cite{retinanet_improv}.
    \vspace{-2mm}
\end{itemize}

\section{Dataset Details}
\label{sec:dataset}
\vspace{-2mm}
DeepLesion~\cite{yan2018deeplesion} is the largest publicly available repository of CT-slices with annotated lesions across different organs of the body released by the National Institutes of Health (NIH). It consists of data from $4,427$ unique patients based on markings performed by radiologists during their routine work. There are approximately $32,120$ axial CT slices from $10,594$ CT studies of the patients having around $1$-$3$ lesions annotated per CT -scan. The lesions in each image have annotations such as bounding-box coordinates and size measurements, etc. which add up to $32,735$ lesions altogether from eight different body organs including bone, abdomen, mediastinum, liver, lung, kidney, soft-tissue, and pelvis. We use the official split of the DeepLesion dataset for training and evaluation of our proposed \emph{DKMA-ULD}.

\vspace{-2mm}
\section{Proposed Method: \emph{DKMA-ULD}}
\label{sec:proposed-method}
\vspace{-2mm}
\emph{DKMA-ULD} method, as shown in Figure~\ref{fig:arch-dia}, consists of different modules such as pre-processing, multi-intensity image generation using five different HU windows, convolutional feature extraction backbone, multi-head self-attention based feature-fusion, lesion-specific anchors, and self-supervision. These modules are discussed in detail as follows:
\begin{itemize}

\item \textbf{Pre-processing}: Typically, clinicians observe multiple adjacent slices of a patient's CT scan to confirm the diagnosis of a pathology. In order to provide $3$D context of a patient's CT-scan to the network, we utilize its $3$ slices (key slice with one superior and one inferior neighboring slice) to generate $3$-channel image. First, we remove black borders of the CT-slices for computational efficiency and to focus on region-of-interest. Next, we normalize and clip the $12$-bit intensity values of a CT slice using a HU window and then, re-scale them to floating-point values in the range $[0.0, 255.0]$. Subsequently, we re-sample all the CT-slices to a common resolution of $0.8 \times 0.8 \times 2$ mm$^3$.
In addition, since the DeepLesion dataset does not have segmentation masks for the lesions, we generate pseudo masks using provided RECIST diameter measurements~\cite{recist}. These pseudo masks boost the performance of Mask-RCNN~\cite{he2017mask} by adding a branch for predicting segmentation masks on region proposals generated by RPN. We also augment the data during training using random affine transformations such as horizontal and vertical flips, resizing with a ratio of $0.8$ to $1.2$, and translation of $(-8,8)$ pixels in $x$ and $y$ direction. 

\begin{figure}[t]
  \centering
  \includegraphics[width=0.65\linewidth]{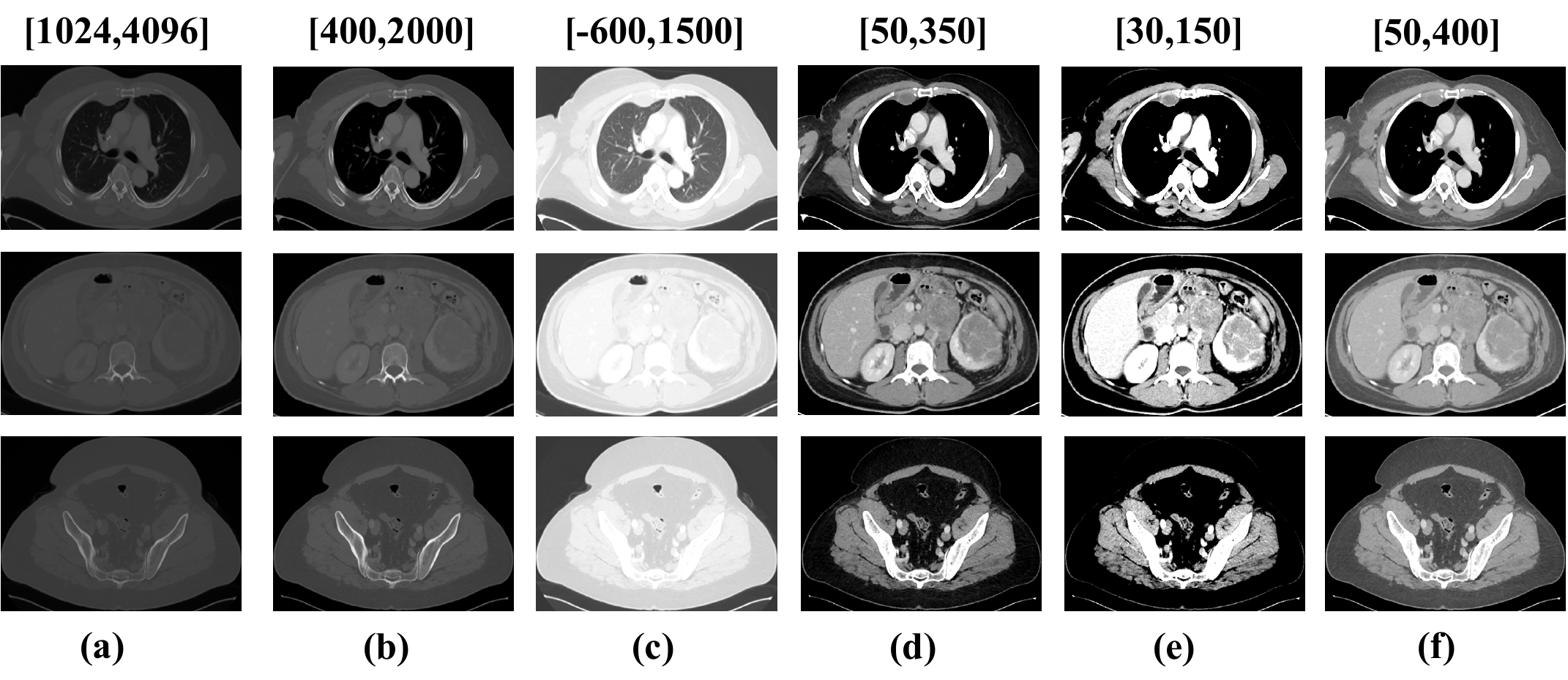}
  \caption{\small{
  The top, middle, and bottom rows have CT-slices of chest-region, abdomen-region, and pelvic-region, respectively. The column (a) illustrates images with commonly used HU window $(U = {[1024,4096]})$ while columns ranging from (b) to (f) have images with our new $5$ HU windows. The figure clearly demonstrates that by using more number of windows, different organs of a particular body-region present in a CT volume, can be highlighted more efficiently.)}}
  \label{fig:hu-windows}
  \vspace{-3mm}
\end{figure}

\item \textbf{Multiple Intensity Image Generation}: In general, the intensity of a CT-slice is re-scaled using a certain HU window, $U$ (e.g., a single and wide window of $[1024, 4096]$) in order to include gray-scale intensities of different organs~\cite{tang2019uldor, 3dce}. However, using a single window suppresses organ-specific information resulting into a degenerated image-contrast, as shown in Figure~\ref{fig:hu-windows}(a), which in turn makes it hard for the network to learn to focus on various organs present in the given CT volume. During manual detection of lesions, radiologists adjust these intensity values to focus on organs/tissue of interest~\cite{bae2005ct}. We exploit this domain knowledge and propose to feed it to the deep network explicitly in the form of CT-slices having multiple intensities which highlight different organs of the body. In a previous method by Zihao Li et al.~\cite{li2019mvp}, a clustering algorithm is used to determine three HU windows. In this paper, we incorporate this multi-organ information in input CT-slices by introducing five novel HU windows which are determined in such a way that the major body organs are covered. The proposed HU windows, as inspired by Masoudi et al.~\cite{masoudi2021quick}, which cover almost all organs of interest for radiologists are: $U_1 = {[400,2000]}$, $U_{2,3} = {[-600,1500], [50,350]}$, $U_4 = {[30,150]}$,  $U_5 = {[50,400]}$ for bones, chest region including lungs \& mediastinum, abdomen including liver \& kidney, and soft-tissues, respectively. For a $U  = [\textbf{U}_{\textbf{l}}, \textbf{U}_{\textbf{w}}]$, where, $\textbf{U}_{\textbf{l}}$ and $\textbf{U}_{\textbf{w}}$ are the window level/center and window width, the intensity values of a CT-slice are first normalized using $\textbf{U}_{\textbf{l}} \pm \textbf{U}_{\textbf{w}}/2$ as data min/max \& clipped between $[0,1]$ and then, re-scaled to values in $[0,255]$.

\vspace{-2mm}
\item \textbf{Convolution Feature Extraction Backbone}: Now, for a given patient, $5$ multiple intensity images each having $3$ slices/channels, are passed as input to the ResNeXt-152 shared backbone with feature pyramid network (FPN)~\cite{lin2017featurefpn} based convolutional feature extractor.
The fact that applying pooling layers of CNN repeatedly on an image can filter out information of small objects due to downsampling, hence, resulting in missing small and medium-sized lesions in radiological scans. Therefore, we utilize FPN where shallow and deeper feature maps are more suitable for detecting small/medium and larger lesions, respectively. As a result, for a given input, we obtain $5$ feature-map blocks ($F_{Ui}$) corresponding to $5$ FPN levels, each having $5$ feature map sub-levels ($P_j$) of dimension $(256,H,W)_j$, where H and W represent height and width of the feature-map and $j$ = $2,...,6$ are the pyramid-levels. These extracted feature maps at different FPN levels, each having a different resolution allows RPN to effectively focus on lesions of different sizes.


\item \textbf{Convolution augmented Multi-head Attention for Feature Fusion}:
Earlier ULD techniques such as MULAN~\cite{yan2019mulan} incorporated information from multiple slices in their network by fusing the feature maps of all 3-channel images/slices with a convolution layer to obtain a 3D-context-enhanced feature map for the central slice. In this work, we propose a novel convolution augmented multi-head attention-based feature fusion module. Recently, vision transformers~\cite{DeiT, ViT, detr} have achieved state-of-the-art results on various machine vision tasks via a focus on self-attention~\cite{attentionisallyouneed}. The use of multi-head self-attention enables the model to attend jointly to both spatial and feature sub-spaces. As shown in Figure~\ref{fig:arch-dia}(b), we first concatenate sub-level feature maps $(256,H,W)_j$ of $5$ different intensities to obtain feature-vectors of shape $(256*5,H,W)_j$. These features are, subsequently, passed to two parallel branches namely, the 2D convolution layer and transformer's multi-head self-attention~\cite{ViT}. Finally, their outputs are fused using concatenation. Since the output depth of the attention module is dependent on the depth of its "values" matrix ($\emph{dv}$), the output depth of $2D$ convolution branch is kept such that the depth of the final feature vector obtained after concatenation of both the outputs is $256$. Similar attention-based feature fusion is used at all $5$ feature-map sub-levels and finally, we obtain a fused feature map block ($F'$, with $5$ feature-maps sub-levels) for later processing. To minimize computation overhead for attention, we use $2$ attention heads (${N_{h}}$) and keep the depth of values matrix as $4$. In addition, we use $20$ dimensions per head for key and query matrix~\cite{aug_attention}.

\item \textbf{Lesion-specific Anchors}: 
Next, for extracting Regions of Interest (ROI) from the obtained feature maps from $5$ FPN levels, anchor boxes play a very crucial role. We observed that the small lesions are hard to detect using default anchor sizes and ratios \cite{he2017mask, lin2017featurefpn} used in RPN for real-world object detection. To circumvent this, we propose new custom anchors which are well suited for detecting lesions of all sizes in CT scans. Let's say, $H$ and $W$ are the image height and width, respectively. We generate anchor boxes of different sizes centered on each image pixel such that it has maximum IoU with lesion bounding box. If anchor sizes and ratios are in sets $\{s_1, s_2, ...,s_n\}$ and $\{r_1, r_2, ...,r_m\}$, respectively for each $r>0$, we will have a total of $WH(n+m-1)$ anchor boxes~\cite{zhang2020dive}. Consider $w_b$ and $h_b$ to be the width and height of anchor boxes,
\vspace{-2mm}
\begin{equation}
    [w_b, h_b] = [{W}{s_n}\sqrt{r_m}, {{H}{s_n}}/{\sqrt{r_m}}] \quad s.t. \quad n,m\in[1,5]
    \vspace{-2mm}
\end{equation}

We employ a differential evolution search algorithm~\cite{storn1997differential} and find $5$ best anchor sizes $[16, 24, 64, 128, 256]$ and ratios $[3.27, 1.78, 1, 0.56, 0.30]$ for $P2, P3, P4, P5, P6$ feature map sub-levels, respectively. These lesion-specific anchors are used in RPN for RoI extraction, which are combined with feature maps using RoI pooling layer and further used, for predicting bounding-boxes around lesions along with probability values, as shown in Figure~\ref{fig:arch-dia}. We demonstrate that the custom anchors allow us to cover varied-sized lesions and more specifically, improve the detection of small-sized ($< 10$mm) and medium-sized ($10-30$mm) lesions considerably, as evident in Figure~\ref{fig:lesion_size}(a). 

\item \textbf{Self supervision}: 
The idea behind self-supervised learning (SSL) is that the learned intermediate representations can carry better semantic and structural meanings and can prove to be beneficial for a variety of downstream tasks. In order to make our \emph{DKMA-ULD} more robust, we utilize a state-of-the-art SSL technique called BYOL~\cite{grill2020bootstrap}. It relies on two neural networks, referred to as online and target networks, that interact and learn from each other. The target network (parameterized by $\xi$) has the same architecture as the online one (parameterized by $\theta$), but with polyak averaged weights, $\xi \leftarrow{\tau \xi + (1 - \tau)\theta}$. The goal is to learn a representation $y$ that can be used in downstream tasks. Generally, the detection networks are initialized using weights pre-trained on Imagenet consisting of natural images and may not be effective for the medical imaging domain. Therefore, we propose domain-specific weights, obtained by training the backbone using SSL over train-split (23K images) of the DeepLesion dataset, for initializing \emph{DKMA-ULD} to obtain enhanced performance.


\end{itemize}

\vspace{-3mm}
\section{Experimental Results and Discussions}
\label{sec:exp-res}
We use the official data split of DeepLesion~\cite{yan2018deeplesion} dataset which consists of 70\%, 15\%, 15\% for training, validation, and test, respectively. Please note that the DeepLesion test-set includes only key CT-slices and may contain missing annotations. The lesion detection is classified as true positive (TP) when the IoU between the predicted and the ground-truth bounding-box is larger than $0.5$. We report average sensitivity, computed at $0.5$, $1$, $2$, and $4$ false-positives (FP) per image, as the evaluation metric on the standard test-set split for the fair comparison.\\

\begin{table}[t]
\begin{center}
\setlength{\tabcolsep}{0.1\tabcolsep}
\begin{tabular}{|l|c|l|c|c|c|c|c|}
\hline
\textbf{Method} & \textbf{Windows}  & \textbf{FP@0.5} & \textbf{FP@1.0} & \textbf{FP@2.0} & \textbf{FP@4.0} & \textbf{Average}
\\\hline
3DCE(27 slices)~\cite{3dce} & 1  & 62.48 & 73.37 & 80.70 & 85.65 & 75.55 \\
improved RetinaNet(3 slices)~\cite{retinanet_improv} & 1  & 72.18 & 80.07 & 86.40 & 90.77 & 82.36 \\
MVP Net(9 slices)~\cite{li2019mvp} & 3 & 73.83 & 81.82 & 87.60 & 91.30 & 83.64 \\
MULAN(27 slices, w/o tags)~\cite{yan2019mulan} & 1  & 76.10 & 82.50 & 87.50 & 90.90 & 84.33 \\
MULAN(27 slices, w/ 171 tags)~\cite{yan2019mulan} & 1  & 76.12 & 83.69 & 88.76 & 92.30 & 85.22 \\ 
MELD(9 slices))~\cite{meld} & 1  & 77.80 & 84.80 & 89.00 & 91.80 & 85.90\\
MELD+MAM+NRM(9 slices)~\cite{meld} & 1  & 78.60 & 85.50 & 89.60 & 92.50 & 86.60 \\ \hline  

(a) \emph{DKMA-ULD}* & 5 & 78.10 & 85.26 & 90.48 & 93.48 & 86.88 \\
(b)\textbf{+Self-Supervision} & \textbf{5} & \textbf{78.75} & \textbf{85.95} & \textbf{90.48} & \textbf{93.48} & \textbf{87.16}\\

\hline
\end{tabular}
\vspace{2mm}
\caption{\small{Comparison of \emph{DKMA-ULD} with previous state-of-the-art ULD methods. Sensitivity(\%) at different false-positives (FP) per sub-volume on the volumetric test-set of DeepLesion~\cite{yan2018deeplesion} dataset. Here, we feed CT-slices, after performing cropping of black-borders during pre-processing, to the network. *training with Imagenet pre-trained weights.}}
\label{tab:comp-sota}
\end{center}
\vspace{-5mm}
\end{table}

\noindent \textbf{Training}: The proposed \emph{DKMA-ULD} is trained on $3$ channel CT images of size $512 \times 512$ with a batch size of $4$ on a single NVIDIA Tesla V100 having $32$GB GPU-memory. We use cross-entropy and smooth $\ell_1$ loss for classification and bounding-box regression, respectively. The model is trained until convergence using SGD optimizer with a learning rate (LR) and decay-factor of $0.02$ and $10$, respectively. 
The SSL model is trained using cross-entropy loss with a batch size of $64$, Adam optimizer~\cite{kingma2014adam}, and LR of $3e-4$ for $300$ epochs.\\

\begin{figure}[t]
  \centering
  \includegraphics[width=0.75\linewidth]{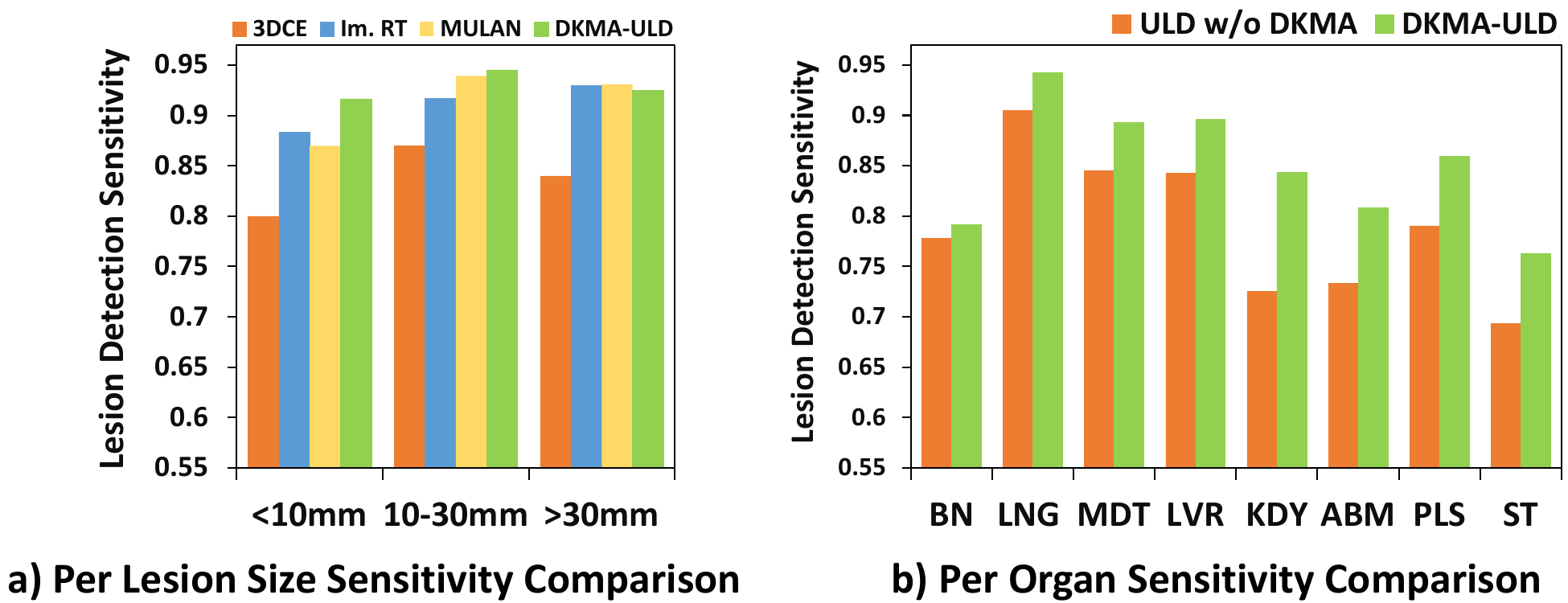}
  \caption{\small{Sensitivity comparison for different lesion sizes and organs. a) Sensitivity (at FP$=4$) for lesions with 3 different size ranges is compared with existing methods~\cite{3dce, retinanet_improv, yan2019mulan} and our proposed DKMA-ULD network w/o self-supervision. b) Average sensitivity comparison per organ computed using our proposed lesion detector without and with domain knowledge \& multi-head attention. Here, BN, LNG, MDT, LVR, KDY, ABM, PLS and ST represent different organs such as bones, lungs, mediastinum, liver, kidney, abdomen, pelvis and soft-tissues, respectively.
  }}
  \label{fig:lesion_size}
  \vspace{-2mm}
\end{figure}

\begin{figure}[t]
  \centering
  \includegraphics[width=0.75\linewidth]{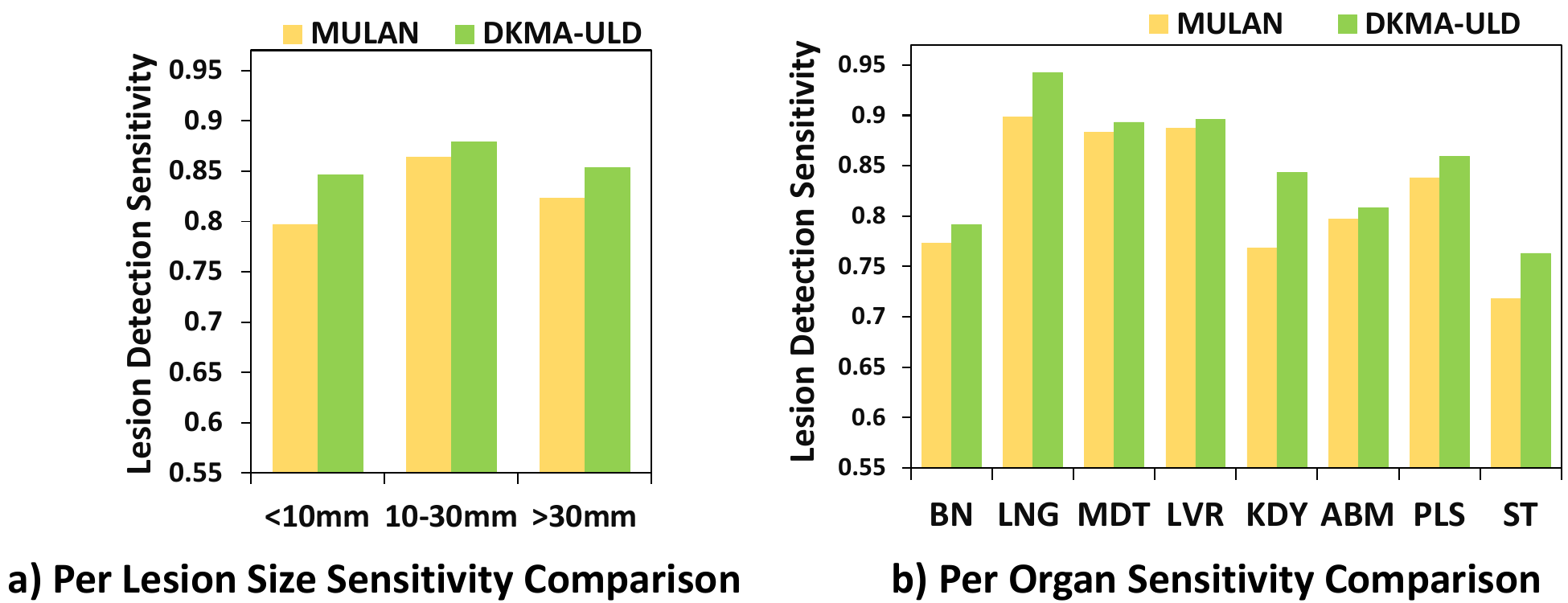}
  \caption{\small{Average sensitivity comparison for different lesion sizes and organs. a) Average sensitivity ($FP = \{0.5, 1, 2, 4\}$) for lesions with $3$ different size ranges is compared with MULAN~\cite{yan2019mulan} and our proposed \emph{DKMA-ULD}. b) Organ-wise average sensitivity of MULAN and DKMA-ULD.
  }}
  \label{fig:lesion_size_mulan}
  \vspace{-2mm}
\end{figure}

\begin{figure}[t]
  \centering
  \includegraphics[width=0.9\linewidth]{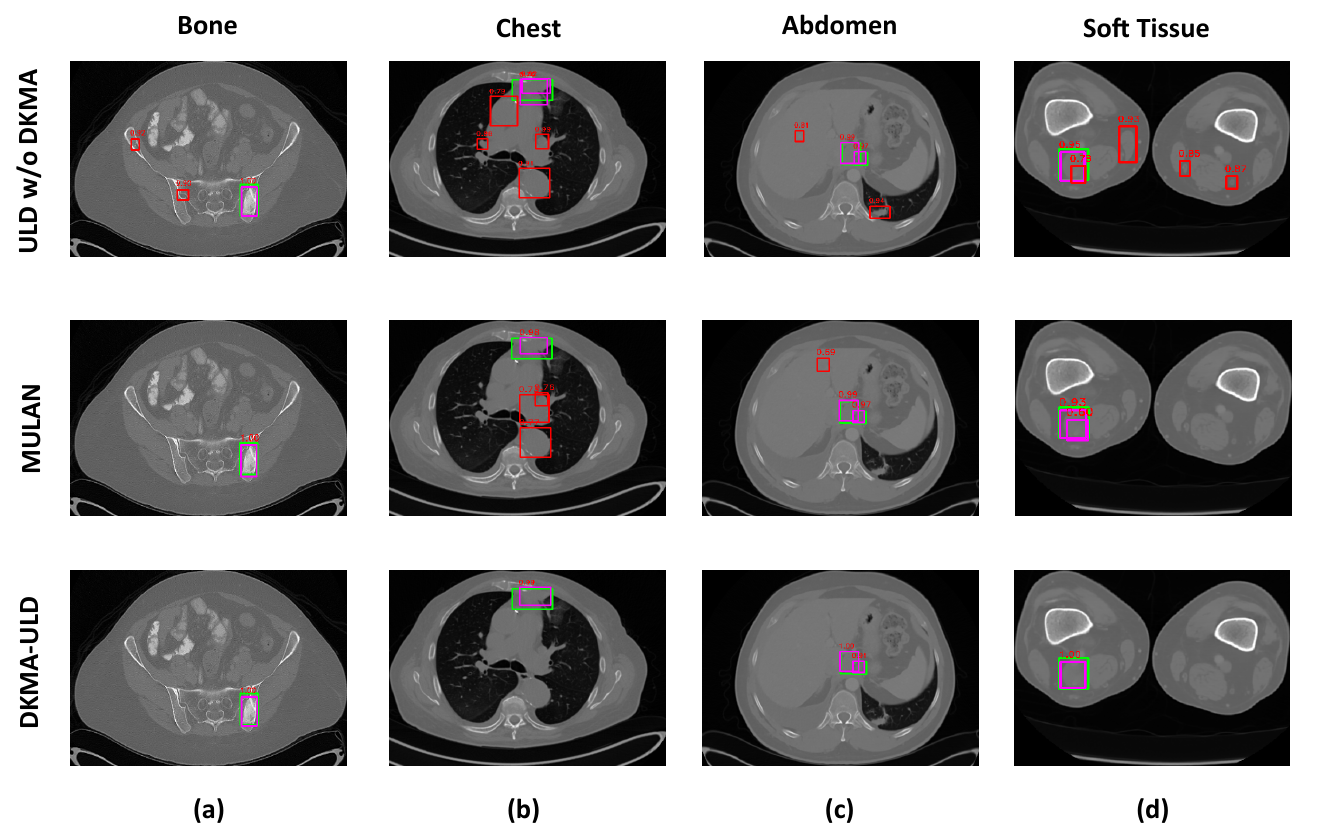}
  \caption{\small{Qualitative comparison of \emph{DKMA-ULD} and MULAN~\cite{yan2019mulan} (at FP  =$2$) on CT-scans of different body regions. 
  The green, magenta, and red color boxes represent ground-truth, true-positive (TP), and false-positive (FP) lesion detection, respectively. Please note that ULD w/o DKMA represents when 3 slices with only one HU window ($[1024, 4096]$), default anchors, and without convolution augmented multi-head attention feature fusion are used. We can observe that after incorporating domain knowledge in the form of multi-intensity CT slices, custom anchors, and multi-head attention (i.e., \emph{DKMA-ULD}), the number of FP reduced drastically resulting in improved lesion detection performance as compared to MULAN.}}
  \vspace{-5mm}
  \label{fig:detection-results}
\end{figure}

\noindent \textbf{Results and Discussions}: We evaluate the performance of our \emph{DKMA-ULD} against previous methods in the literature, as shown in Table~\ref{tab:comp-sota}. Our experiments demonstrate that by using only $3$ slices per patient, the proposed method \emph{DKMA-ULD} outperforms all the previous state-of-the-art ULD methods at different FP per image and achieves an average sensitivity of $86.88\%$ when Imagenet pre-trained weights are used for the backbone initialization. 
Here, one important point to note is that the base model of MELD~\cite{meld} (refer row 6 of Table~\ref{tab:comp-sota}) achieves an average sensitivity of $85.90\%$ by training on $4$ heterogeneous datasets, 
namely LUNA~\cite{luna}, LiTS~\cite{lits}, NIH-Lymph~\cite{nih_lymph}
and 
DeepLesion~\cite{yan2018deeplesion}. 
On the other hand, our proposed \emph{DKMA-ULD} base model w/o self-supervision, despite being trained only on DeepLesion train-set, beats MELD on Deeplesion test-set by achieving an average sensitivity of $86.88\%$. Therefore, it is evident that the introduction of domain-specific features and multi-head attention-based feature fusion have enabled \emph{DKMA-ULD} to learn robust representations. Hence, it validates our claim that domain knowledge can alleviate the requirement of a set of diverse datasets for learning good representations in medical imaging analysis. Furthermore, we observe that previous methods such as MULAN~\cite{yan2019mulan} and MELD~\cite{meld} improved the performance of their base models by incorporating techniques such as the use of tagging in MULAN; Missing Annotation Matching (MAM) and Negative Region Mining (NRM) in MELD (refer row 5 and 7 of Table~\ref{tab:comp-sota}). Hence, we also experimented by initializing our network with self-supervised weights. As shown in Table~\ref{tab:comp-sota}, it leads to a gain in performance and achieves a final average sensitivity of $87.16\%$. For more results on organ-wise sensitivity, refer supplementary material.


Next, we show a comparison of sensitivity at $FP = 4$ for different lesion sizes and average sensitivity (over $FP = \{0.5, 1, 2, 4\}$) for different organs. We observe from the Figure~\ref{fig:lesion_size}(a) that \emph{DKMA-ULD} improves the detection of very small ($<10$mm) and medium-sized ($10-30$mm) lesions over 3DCE~\cite{3dce}, improved RetinaNet~\cite{retinanet_improv}, and MULAN~\cite{yan2019mulan}. Unlike 3DCE~\cite{3dce} and improved RetinaNet~\cite{retinanet_improv}, sensitivity values of MULAN~\cite{yan2019mulan} for different lesion sizes and organs are not mentioned in their paper. Moreover, as the best trained model of MULAN with tags is not available publicly, we use the official released base model of MULAN~\cite{yan2019mulan} (average sensitivity of $84.33\%$) for computing lesion size-wise and organ-wise sensitivity values on DeepLesion test-set. For a fair comparison, we use the base model of our proposed DKMA-ULD without self-supervision (average sensitivity of $86.88\%$) in Figure~\ref{fig:lesion_size} and Figure~\ref{fig:lesion_size_mulan}.  Further, in Figure~\ref{fig:lesion_size}(b), we observe that our proposed method of including domain-specific information in the lesion detection network improves the average sensitivity across all organs. Furthermore, we provide a comparison of average sensitivity of MULAN and DKMA-ULD for different lesion-sizes and organs in Figure~\ref{fig:lesion_size_mulan} and demonstrate that DKMA-ULD substantially improves over MULAN in all the cases. 
For all the above experiments, we use cropped CT slices by clipping the black border region, as mentioned in previous state-of-the-art methods~\cite{yan2019mulan, meld}, to focus on the region of interest. 

\begin{table}[t]
\begin{center}

\setlength{\tabcolsep}{0.6\tabcolsep}
\begin{tabular}{|c|c|l|c|c|c|}
\hline
\textbf{Sr. No.} &\textbf{HU windows} & \textbf{Backbone} &\textbf{Attention} & \textbf{Custom Anchors} & \textbf{Avg. Sensitivity} 
\\ \hline
1 &1 & x101 &  &  & 77.59 \\ 
2 &3 & x101 &  &  & 80.66 \\
3 &5 & x101 &  &  & 82.37 \\
4 & 5 & x101 & \checkmark &  & 83.30 \\ 
5 & 5 & x101 & \checkmark & \checkmark & 84.23 \\ 
6 & 5 & x152 & \checkmark & \checkmark & 84.85 \\ 
\textbf{ 7*} &\textbf{5} & \textbf{x152} & \textbf{\checkmark} & \textbf{\checkmark} & \textbf{86.88} \\ 
\hline
\end{tabular}
\vspace{2mm}
\caption{\small{Ablation studies and average sensitivity comparison (\%) of introducing different modules in the proposed lesion detection (\emph{DKMA-ULD}) on the test-set of the DeepLesion dataset. *CT-slices after cropping of outer black-region are used for this experiment.}}
\label{tab:ablation}

\end{center}
\vspace{-10mm}
\end{table}

Now, we present the ablation study on the introduction of different modules in the proposed lesion detection pipeline, as shown in Table~\ref{tab:ablation}. Our proposed $5$ HU windows to give organ-specific domain knowledge results in an improvement of approximately $2\%$ in the average sensitivity ($82.37\%$), as shown in row 3 of Table~\ref{tab:ablation}. Subsequent to this, we experiment with the inclusion of our novel convolution augmented multi-head attention module for feature fusion and custom anchors to detect varied-sized lesions effectively. We observe a performance boost by achieving an average sensitivity of $84.23\%$. All the ablation experiments are performed on CT-slices without applying cropping during the pre-processing step. Later in our experiments, we replace our feature extraction backbone with ResNeXt-152 and clip black borders in CT slices enabling the network to focus only on the region of interest. This resulted in a quantitative improvement by achieving a state-of-the-art average sensitivity of $86.88\%$. Finally, we  show a qualitative comparison of lesion detection performance of our proposed \emph{DKMA-ULD} in the form of reduction of FP in Figure~\ref{fig:detection-results}. For more detailed experimental results, please refer supplementary material.

\vspace{-2mm}
\section{Conclusion and Future Work}
\label{sec:conclusion}

In this paper, we demonstrate the potential of exploiting domain knowledge in medical imaging data for developing a robust universal lesion detection network named \emph{DKMA-ULD} that detects lesions across multiple organs of interest with improved sensitivity as compared to the state-of-the-art methods. We also prove that domain-specific weight initialization of ULD using self-supervision on the DeepLesion dataset gives a boost in lesion detection performance. Further, we provide evidence for our choices in using multiple HU windows, lesion-specific custom anchors, multi-head attention-based feature fusion and surpass the state-of-the-art ULD methods by achieving a sensitivity of $87.16\%$ with only $3$ slices per patient. This idea of exploiting maximum information in the dataset and learning self-supervised weights can prove useful for any medical image analysis task. 
In the future, we intend to extend this work by developing an anchor-less lesion detection network and making it robust to shifts in the domain, acquisition protocols, etc. using domain adaptation techniques. 


\bibliography{egbib}
\end{document}


\maketitle

\section{Ablation for Feature Extraction Backbone}
\let\thefootnote\relax\footnote{* Equal contribution in paper}
We use $5$ multiple intensity images having $3$ slices/channels each for a given patient in DeepLesion~\cite{yan2018deeplesion} dataset. These images are passed as input to the shared convolutional feature extractor with feature pyramid network (FPN)~\cite{lin2017featurefpn}. To determine the best feature extractor network that can learn the most relevant features for 3D CT-scans, we performed experiments with different ResNet~\cite{he2016deep} variants. We found that ResNeXt-152 performs the best, as it is evident in Table~\ref{tab:ablation_resnet}. Therefore, in our proposed \emph{DKMA-ULD} method, we utilize the ResNeXt-152 backbone for feature extraction with self-supervised weights using BYOL technique~\cite{grill2020bootstrap} and achieve state-of-the-art average sensitivity of $87.16\%$ on the test-set of the DeepLesion dataset.



\begin{table}[!ht]
\centering
\resizebox{\textwidth}{!}{%
\begin{tabular}{|l|l|c|c|c|c|c|}
\hline
\textbf{Model} &
  \textbf{Backbone} &
  \multicolumn{1}{l|}{\textbf{FP@0.5}} &
  \multicolumn{1}{l|}{\textbf{FP@1.0}} &
  \multicolumn{1}{l|}{\textbf{FP@2.0}} &
  \multicolumn{1}{l|}{\textbf{FP@4.0}} &
  \multicolumn{1}{l|}{\textbf{Average}} \\ \hline
\textbf{DKMA-ULD}        & \textbf{x101} & 75.09          & 83.88          & 89.28          & 92.83          & 85.27          \\ \hline
\textbf{DKMA-ULD + BYOL} & \textbf{x101} & 76.07          & 84.31          & 89.44          & 92.94          & 85.69          \\ \hline
\textbf{DKMA-ULD}        & \textbf{x152} & 78.10          & 85.26          & 90.48          & 93.48          & 86.88          \\ \hline
\textbf{DKMA-ULD + BYOL} & \textbf{x152} & \textbf{78.75} & \textbf{85.95} & \textbf{90.48} & \textbf{93.48} & \textbf{87.16} \\ \hline
\end{tabular}%
\label{tab:ablation_resnet}
}

\vspace{2mm}
\caption{Sensitivity(\%) for DKMA-ULD using different backbone networks and weight initialization via self-supervised learning (SSL), at different false-positives (FP) per sub-volume on the volumetric test-set of DeepLesion~\cite{yan2018deeplesion} dataset.}
\end{table}




\section{Organ-wise sensitivity}
DeepLesion dataset~\cite{yan2018deeplesion} consists of approx. $32K$ annotated lesions across $8$ different organs of the body. It is the largest dataset available right now which contains lesions in a variety of organs and hence, the best candidate for developing a universal lesion detection network. Our proposed method \emph{DKMA-ULD} uses multiple-HU windows and lesion-specific custom anchors with a novel multi-head self attention-based feature fusion module. The inclusion of domain-knowledge specific features in the proposed \emph{DKMA-ULD} has led to an improved overall and organ-wise performance as evident in Table~\ref{tab:Per-Organ sensitivity}.

\begin{table}[!ht]
\centering

\begin{tabular}{|l|c|c|c|c|c|}
\hline
\textbf{Organ Type} &
  \multicolumn{1}{l|}{\textbf{FP@0.5}} &
  \multicolumn{1}{l|}{\textbf{FP@1.0}} &
  \multicolumn{1}{l|}{\textbf{FP@2.0}} &
  \multicolumn{1}{l|}{\textbf{FP@0.5}} &
  \multicolumn{1}{l|}{\textbf{Average}} \\ \hline 
\textbf{Bone}        & 65.74 & 79.63 & 84.26 & 87.03 & 79.16 \\ \hline
\textbf{Lung}        & 89.45 & 93.27 & 95.72 & 96.45 & 93.72 \\ \hline
\textbf{Mediastinum} & 84.34 & 90.37 & 93.27 & 95.35 & 90.83 \\ \hline
\textbf{Liver}       & 83.00 & 89.00 & 93.42 & 95.57 & 90.24 \\ \hline
\textbf{Kidney}      & 77.58 & 82.75 & 91.37 & 93.10 & 86.20  \\ \hline
\textbf{Abdomen}     & 67.88 & 78.96 & 85.52 & 90.12 & 80.62 \\ \hline
\textbf{Pelvis}      & 76.83 & 83.41 & 88.29 & 91.71 & 85.06 \\ \hline
\textbf{Soft Tissue} & 62.75 & 70.96 & 78.59 & 84.16 & 74.11 \\ \hline
\end{tabular}%
\caption{Organ-wise sensitivity in $\%$ (at different FP and average) for \emph{DKMA-ULD} having BYOL initialized ResNeXt-152 backbone.}

\label{tab:Per-Organ sensitivity}
\end{table}

\begin{table}[!ht]
\centering

\resizebox{\textwidth}{!}{%
\begin{tabular}{|l|c|c|c|c|c|c|c|c|}
\hline
\textbf{Method} & \textbf{BN} & \textbf{LNG} & \textbf{MDT} & \textbf{LVR} & \textbf{KDY} & \textbf{ABM} & \textbf{PLS} & \textbf{ST} \\ \hline
MULAN (w/o tags)~\cite{yan2019mulan}  & 77.31 & 89.86 &  88.37 & 88.75 & 76.83 & 79.69 & 83.84 & 71.80  \\
DKMA-ULD(ours w/o byol) & \textbf{79.16} & \textbf{94.30} & \textbf{89.32} & \textbf{89.64} & \textbf{84.37} & \textbf{80.81} & \textbf{85.97} & \textbf{76.31} 
\\ \hline

\end{tabular}%
}
\vspace{2mm}
\caption{Organ-wise average sensitivity (\%) comparison (over $FP = \{0.5, 1, 2, 4$\}) of base MULAN~\cite{yan2019mulan} model w/o tags and our proposed base \emph{DKMA-ULD} model (without byol self-supervision) on test-set of DeepLesion~\cite{yan2018deeplesion} dataset.}
\label{tab:comp-mulan}
\end{table}

\begin{figure}[]
  \centering
  \includegraphics[width=0.75\linewidth]{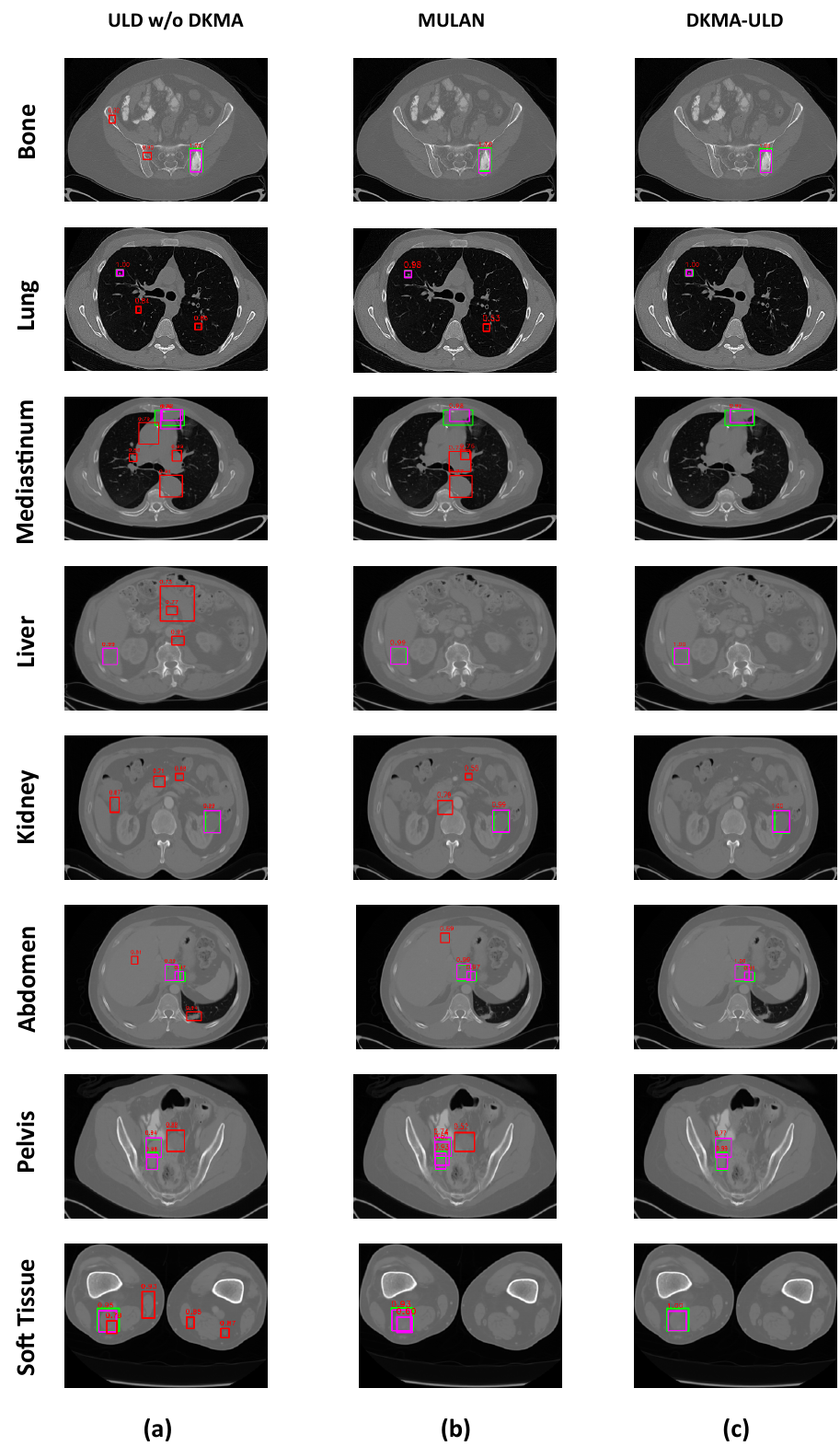}
  \caption{\small{Qualitative comparison of \emph{DKMA-ULD} and MULAN~\cite{yan2019mulan} (at FP  =$2$) on CT-scans of different body regions. 
  The green, magenta, and red color boxes represent ground-truth, true-positive (TP), and false-positive (FP) lesion detection, respectively. Please note that ULD w/o DKMA represents when 3 slices with only one HU window ($[1024, 4096]$), default anchors, and without convolution augmented multi-head attention feature fusion are used. We can observe that after incorporating domain knowledge in the form of multi-intensity CT slices, custom anchors, and multi-head attention (i.e., \emph{DKMA-ULD}), the number of FP reduced drastically resulting in improved lesion detection performance as compared to MULAN.}}
  \label{fig:oran_detection-results}
  \vspace{-2mm}
\end{figure}

Next, we provide detailed comparison of our proposed DKMA-ULD network with MULAN~\cite{yan2019mulan}. Since, the best trained model of MULAN with tags is not available publicly, we use the official released base model of MULAN~\cite{yan2019mulan} for computing organ-wise sensitivity values on DeepLesion test-set for comparison. For a fair comparison, we use the base model of our proposed DKMA-ULD without self-supervision in Table~\ref{tab:comp-mulan}. It is clearly visible that our proposed method DKMA-ULD outperforms MULAN in lesion detection across all organs. Further, We present a qualitative comparison of our proposed DKMA-ULD with MULAN~\cite{yan2019mulan} in Figure~\ref{fig:oran_detection-results} and demonstrate that the detection of false positives is substantially reduced using domain-knowledge.

\bibliography{sup_egbib}